\documentclass[11pt,a4paper]{article}
\usepackage[hyperref]{emnlp-ijcnlp-2019}
\usepackage{times}
\usepackage{latexsym}

\usepackage{url}
\usepackage{booktabs}
\usepackage{times}
\usepackage{latexsym}
\usepackage{hyperref}       
\usepackage{url}            
\usepackage{amsfonts, amssymb}       
\usepackage{nicefrac}       
\usepackage{microtype}      
\usepackage{graphicx}
\usepackage{capt-of}
\usepackage{float}
\usepackage{mathtools}
\usepackage{array}
\usepackage{url}
\usepackage{amsmath}

\aclfinalcopy 

\title{Taskmaster-1: Toward a Realistic and Diverse Dialog Dataset }

\author{
    \textbf{
        Bill Byrne$^1$\thanks{~~Equal Contribution} \quad
        Karthik Krishnamoorthi$^1$\footnotemark[1] \quad
        Chinnadhurai Sankar$^2$\footnotemark[1] \quad
        Arvind Neelakantan$^1$
    } \\
    \textbf{
        Daniel Duckworth$^1$ \quad
        Semih Yavuz$^3$ \quad
        Ben Goodrich$^1$ 
    } \\
    \textbf{
        Amit Dubey$^1$    \quad
        Andy Cedilnik$^1$ \quad
        Kyu-Young Kim$^1$
    } \\
  $^1$Google LLC, Mountain View, California \\
  $^2$Mila, Universit\'e de Montr\'eal \\
  $^3$University of California, Santa Barbara \\
    {\tt \{billb,krishnamoorthi,aneelakantan\}@google.com}\\
}

\date{}

\begin{document}
\maketitle
\begin{abstract}
 A significant barrier to progress in data-driven approaches to building dialog systems is the lack of high quality, goal-oriented conversational data. To help satisfy this elementary requirement, we introduce the initial release of the Taskmaster-1 dataset which includes 13,215 task-based dialogs comprising six domains. Two procedures were used to create this collection, each with unique advantages. The first involves a two-person, spoken ``Wizard of Oz" (WOz) approach in which trained agents and crowdsourced workers interact to complete the task while the second is ``self-dialog" in which crowdsourced workers write the entire dialog themselves. We do not restrict the workers to detailed scripts or to a small knowledge base and hence we observe that our dataset contains more realistic and diverse conversations in comparison to existing datasets. We offer several baseline models including state of the art neural seq2seq architectures with benchmark performance as well as qualitative human evaluations. Dialogs are labeled with API calls and arguments, a simple and cost effective approach which avoids the requirement of complex annotation schema. The layer of abstraction between the dialog model and the service provider API allows for a given model to interact with multiple services that provide similar functionally. Finally, the dataset will evoke interest in written vs. spoken language, discourse patterns, error handling and other linguistic phenomena related to dialog system research, development and design.\footnote{Dataset available at https://g.co/dataset/taskmaster-1}
\end{abstract}

\section{Introduction}
Voice-based ``personal assistants" such as Apple's SIRI, Microsoft's Cortana, Amazon Alexa, and the Google Assistant have finally entered the mainstream. This development is generally attributed to major breakthroughs in speech recognition and text-to-speech (TTS) technologies aided by recent progress in deep learning \cite{deep}, exponential gains in compute power \cite{gpu,tpu}, and the ubiquity of powerful mobile devices. The accuracy of machine learned speech recognizers \cite{asr} and speech synthesizers \cite{wavenet} are good enough to be deployed in real-world products and this progress has been driven by publicly available labeled datasets. However, conspicuously absent from this list is equal progress in machine learned conversational natural language understanding (NLU) and generation (NLG). The NLU and NLG components of dialog systems starting from the early research work \cite{eliza} to the present commercially available personal assistants largely rely on rule-based systems. The NLU and NLG systems are often carefully programmed for very narrow and specific cases \cite{aog,alexa}. General understanding of natural spoken behaviors across multiple dialog turns, even in single task-oriented situations, is by most accounts still a long way off. In this way, most of these products are very much hand crafted, with inherent constraints on what users can say, how the system responds and the order in which the various subtasks can be completed. They are high precision but relatively low coverage. Not only are such systems unscalable, but they lack the flexibility to engage in truly natural conversation.

Yet none of this is surprising. Natural language is heavily context dependent and often ambiguous, especially in multi-turn conversations across multiple topics. It is full of subtle discourse cues and pragmatic signals whose patterns have yet to be thoroughly understood. Enabling an automated system to hold a coherent task-based conversation with a human remains one of computer science's most complex and intriguing unsolved problems \cite{eliza}. In contrast to more traditional NLP efforts, interest in statistical approaches to dialog understanding and generation aided by machine learning has grown considerably in the last couple of years \cite{pipeline,e2egoal,dst}. However, the dearth of high quality, goal-oriented dialog data is considered a major hindrance to more significant progress in this area \cite{e2egoal,ubuntu}. 

To help solve the data problem we present Taskmaster-1, a dataset consisting of 13,215 dialogs, including 5,507 spoken and 7,708 written dialogs created with two distinct procedures. Each conversation falls into one of six domains: ordering pizza, creating auto repair appointments, setting up ride service, ordering movie tickets, ordering coffee drinks and making restaurant reservations. For the spoken dialogs, we created a “Wizard of Oz” (WOz) system \cite{woz} to collect two-person, spoken conversations. Crowdsourced workers playing the ``user" interacted with human operators playing the “digital assistant” using a web-based interface. In this way, users were led to believe they were interacting with an automated system while it was in fact a human, allowing them to express their turns in natural ways but in the context of an automated interface. We refer to this spoken dialog type as ``two-person dialogs". For the written dialogs, we engaged crowdsourced workers to write the full conversation themselves based on scenarios outlined for each task, thereby playing roles of both the user and assistant. We refer to this written dialog type as ``self-dialogs". In a departure from traditional annotation techniques \cite{dst,pipeline,multiwoz}, dialogs are labeled with simple API calls and arguments. This technique is much easier for annotators to learn and simpler to apply. As such it is more cost effective and, in addition, the same model can be used for multiple service providers.

Taskmaster-1 has richer and more diverse language than the current popular benchmark in task-oriented dialog, MultiWOZ \cite{multiwoz}. Table \ref{table1: one_person__multiwoz} shows that Taskmaster-1 has more unique words and is more difficult for language models to fit. We also find that Taskmaster-1 is more realistic than MultiWOZ. Specifically, the two-person dialogs in Taskmaster-1 involve more real-word entities than seen in MutliWOZ since we do not restrict conversations to a small knowledge base. Beyond the corpus and the methodologies used to create it, we present several baseline models including state-of-the-art neural seq2seq architectures together with perplexity and BLEU scores. We also provide qualitative human performance evaluations for these models and find that automatic evaluation metrics correlate well with human judgments. We will publicly release our corpus containing conversations, API call and argument annotations, and also the human judgments.   

\begin{table}

\begin{tabular}{lcc}
\toprule
Statistic                & Self-dialogs & MultiWOZ \\ 
\midrule
\# unique words         & \textbf{21,894}     & 19,175 \\
\# unique named         & \textbf{8,218}     & 1,338 \\
entities & &\\
\# utterances            & \textbf{169,469}           & 132,610         \\
\# dialogs               & 7,708           & \textbf{10,438}          \\
Avg. utterances          & \textbf{21.99}   & 13.70          \\
per dialog & & \\
Avg. tokens              & 8.62             & \textbf{13.82}          \\
per utterance & & \\
Perplexity               & \textbf{17.08}            & 15.62          \\
BLEU                     & \textbf{6.53}             & 11.02          \\ \bottomrule

\end{tabular}
\caption{Statistics comparison: Self-dialogs vs 
MultiWOZ corpus both containing approximately 10k dialogues each.}
\label{table1: one_person__multiwoz}
\end{table}

\section{Related work}

\subsection{Human-machine vs. human-human dialog}
 \citet{survey} discuss the major features and differences among the existing offerings in an exhaustive and detailed survey of available corpora for data driven learning of dialog systems. One important distinction covered is that of human-human vs. human-machine dialog data, each having its advantages and disadvantages. Many of the existing task-based datasets have been generated from deployed dialog systems such as the Let’s Go Bus Information System \cite{go} and the various Dialog State Tracking Challenges (DSTCs) \cite{dst-1}. However, it is doubtful that new data-driven systems built with this type of corpus would show much improvement since they would be biased by the existing system and likely mimic its limitations \cite{partial}. Since the ultimate goal is to be able to handle complex human language behaviors, it would seem that human-human conversational data is the better choice for spoken dialog system development \cite{multiwoz}. However, learning from purely human-human based corpora presents challenges of its own. In particular, human conversation has a different distribution of understanding errors and exhibits turn-taking idiosyncrasies which may not be well suited for interaction with a dialog system \cite{partial,survey}. 

\subsection{The Wizard of Oz (WOz) Approach and MultiWOZ}
The WOz framework, first introduced by \citet{woz} as a methodology for iterative design of natural language interfaces, presents a more effective approach to human-human dialog collection. In this setup, users are led to believe they are interacting with an automated assistant but in fact it is a human behind the scenes that controls the system responses. Given the human-level natural language understanding, users quickly realize they can comfortably and naturally express their intent rather than having to modify behaviors as is normally the case with a fully automated assistant. At the same time, the machine-oriented context of the interaction, i.e. the use of TTS and slower turn taking cadence, prevents the conversation from becoming fully fledged, overly complex human discourse. This creates an idealized spoken environment, revealing how users would openly and candidly express themselves with an automated assistant that provided superior natural language understanding.

Perhaps the most relevant work to consider here is the recently released MultiWOZ dataset \cite{multiwoz}, since it is  similar in size, content and collection methodologies. MultiWOZ has roughly 10,000 dialogs which feature  several domains and topics. The dialogs are annotated with both dialog states and dialog acts. MultiWOZ is an entirely written corpus and uses crowdsourced workers for both assistant and user roles. In contrast, Taskmaster-1 has roughly 13,000 dialogs spanning six domains and annotated with API arguments. The two-person spoken dialogs in Taskmaster-1 use crowdsourcing for the user role but trained agents for the assistant role. The assistant's speech is played to the user via TTS. The remaining 7,708 conversations in Taskmaster-1 are self-dialogs, in which crowdsourced workers write the entire conversation themselves. As \citet{edina,self} show, self dialogs are surprisingly rich in content.

\begin{figure}
\small
\begin{tabular}{rp{5.5cm}}
ASSISTANT: &How can I help you?\\
USER: &Hi, could you help me with booking movie tickets for tonight?\\
ASSISTANT: &What movie are you interested in?\\
USER: &The Upside.\\
ASSISTANT: &Did you have a theater in mind?\\
USER: &Could you check if the Regal Neshaminy... No, AMC Neshaminy in Neshaminy, PA is playing it?\\
ASSISTANT: &Could you spell that?\\
USER: &Sure, n e s h a m i n y.\\
ASSISTANT: &I have a showtime at 7:30 and at 10:30, is that okay?\\
USER: &Yes, could you get two tickets for the 7:30?\\
ASSISTANT: &One moment. Okay so that's 2 tickets for 7:30 at the AMC Neshaminy 24?\\
USER: &Yes.\\
ASSISTANT: &It'll be twenty-four ninety-nine for your tickets.\\
USER: &That sounds great.\\
ASSISTANT: &I've confirmed your tickets, they'll arrive via text shortly. Did you need any other information?\\
USER: &No, that was it. Thank you so much for your help.\\
ASSISTANT: &Great, no problem. I hope you have fun.\\
USER: &I hope so, too. Thank you so much.\\
\end{tabular}
\caption{Sample Taskmaster-1 two-person dialog}
\label{fig:example}
\end{figure}

\begin{figure}
\small

MAIN TASK: Users will pretend they are using a voice-powered personal digital assistant to book movie tickets for a film they ALREADY have in mind.
\begin{enumerate}
\item In several turns (not just one!), cover the following:
    \begin{enumerate}
    \item Film name
    \item Number of people
    \item City
    \item Theater
    \item Time
    \item If applicable: 3D vs. IMAX vs. standard.
    \end{enumerate}

\item They may also want to know things like:
    \begin{enumerate}
      \item Run time
      \item End time
      \item Director, actors, etc.
    \end{enumerate}

\item Make sure to CONFIRM all the relevant ticket details before the end of the dialogue INCLUDING:
    \begin{enumerate}
      \item Total cost for two tickets
      \item Time, location, theater
    \end{enumerate}
    
\item You can assume you have the user’s account info with the ticket service--so no credit card information is necessary.
\item After confirming the details, end the conversation by confirming that the tickets are being sent to the user’s mobile device as a text message. 
\end{enumerate}
\caption {Sample instructions for agents playing ``assistant" role}
\label{fig:agent_ins}
\end{figure}

\begin{figure}
\small

MAIN TASK: Pretend you are using your voice-powered digital assistant to book movie tickets.
 
\begin{enumerate}
\item Start by thinking of a particular movie PLAYING NOW in theaters that you'd like to see. (Use the internet to find one if necessary.)

\item Choose a DIFFERENT CITY from where you live, work, or happen to be at the moment.

\item Pretend you've decided to see this movie tonight and you're taking a friend.

\item The assistant will ask about all relevant details BUT you should make sure it covers all your needs.

\item You can assume you already have an account with the ticket service--so no credit card information is necessary.

\item The assistant will end the conversation by confirming that your tickets are being sent to your mobile device as a text message. (And you can respond thanks, goodbye, ok, etc. for a final closing turn, if you like).

\end{enumerate}

\caption {Sample instructions for crowdsourced workers playing ``user" role}
\label{fig:user_ins}
\end{figure}

\section{The Taskmaster Corpus}
\subsection{Overview}
There are several key attributes that make Taskmaster-1 both unique and effective for data-driven approaches to building dialog systems and for other research.

\vspace{1em}
\noindent {\bf Spoken and written dialogs:} While the spoken sources more closely reflect conversational language \cite{spoken-written}, written dialogs are significantly cheaper and easier to gather. This allows for a significant increase in the size of the corpus and in speaker diversity.

\vspace{.5em}
\noindent {\bf Goal-oriented dialogs:} All dialogs are based on one of six tasks: ordering pizza, creating auto repair appointments, setting up rides for hire, ordering movie tickets, ordering coffee drinks and making restaurant reservations.

\vspace{.5em}
\noindent {\bf Two collection methods:} The two-person dialogs and self-dialogs each have pros and cons, revealing interesting contrasts.

\vspace{.5em}
\noindent {\bf Multiple turns:} The average number of utterances per dialog is about 23 which ensures context-rich language behaviors.

\vspace{.5em}
\noindent {\bf API-based annotation:} The dataset uses a simple annotation schema providing sufficient grounding for the data while making it easy for workers to apply labels consistently.

\vspace{.5em}
\noindent {\bf Size:} The total of 13,215 dialogs in this corpus is on par with similar, recently released datasets such as MultiWOZ \cite{multiwoz}.

\subsection{Two-person, spoken dataset}
\label{sec:woz}
In order to replicate a two-participant, automated digital assistant experience, we built a WOz platform that pairs agents playing the digital assistant with crowdsourced workers playing the user in task-based conversational scenarios. An example dialog from this dataset is given in Figure \ref{fig:example}.

\subsubsection{WOz platform and data pipeline}
While it is beyond the scope of this work to describe the entire system in detail, there are several platform features that help illustrate how the process works.

\vspace{.5em}
\noindent {\bf Modality:} The agents playing the assistant type their input which is in turn played to the user via text-to-speech (TTS) while the crowdsourced workers playing the user speak aloud to the assistant using their laptop and microphone. We use WebRTC to establish the audio channel. This setup creates a digital assistant-like communication style.

\vspace{.5em}
\noindent {\bf Conversation and user quality control:} Once the task is completed, the agents tag each conversation as either successful or problematic depending on whether the session had technical glitches or user behavioral issues. We are also then able to root out problematic users based on this logging.

\vspace{.5em}
\noindent {\bf Agent quality control:} Agents are required to login to the system which allows us to monitor performance including the number and length of each session as well as their averages.

\vspace{.5em}
\noindent {\bf User queuing:} When there are more users trying to connect to the system than available agents, a queuing mechanism indicates their place in line and connects them automatically once they move to the front of the queue.

\vspace{.5em}
\noindent {\bf Transcription:} Once complete, the user's audio-only portion of the dialog is transcribed by a second set of workers and then merged with the assistant's typed input to create a full text version of the dialog. Finally, these conversations are checked for transcription errors and typos and then annotated, as described in Section \ref{sec:annotation}.

\subsubsection{Agents, workers and training}
\label{section:agents}
Both agents and crowdsourced workers are given written instructions prior to the session. Examples of each are given in Figure \ref{fig:agent_ins} and Figure \ref{fig:user_ins}. The instructions continue to be displayed on screen to the crowdsourced workers while they interact with the assistant. Instructions are modified at times (for either participant or both) to ensure broader coverage of dialog scenarios that are likely to occur in actual user-assistant interactions. For example, in one case users were asked to change their mind after ordering their first item and in another agents were instructed to tell users that a given item was not available. Finally, in their instructions, crowdsourced workers playing the user are told they will be engaging in conversation with “a digital assistant”. However, it is plausible that some suspect human intervention due to the advanced level of natural language understanding from the assistant side.

Agents playing the assistant role were hired from a pool of dialog analysts and given two hours of training on the system interface as well as on how to handle specific scenarios such as uncooperative users and technical glitches. Uncooperative users typically involve those who either ignored agent input or who rushed through the conversation with short phrases. Technical issues involved dropped sessions (e.g. WebRTC connections failed) or cases in which the user could not hear the agent or vice-versa. In addition, weekly meetings were held with the agents to answer questions and gather feedback on their experiences. Agents typically work four hours per day with dialog types changing every hour. Crowdsourced workers playing the user are accessed using Amazon Mechanical Turk. Payment for a completed dialog session lasting roughly five to seven minutes was typically in the range of $\$1.00$ to $\$1.30$. Problematic users are detected either by the agent involved in the specific dialog or by post-session assessment and removed from future requests.

\begin{figure}[!ht]
\small
\begin{enumerate}
\item Think of a particular movie PLAYING NOW in theaters that you'd like to see. (Use the internet to find one if necessary.)
\item Choose a DIFFERENT CITY from where you live, work, or happen to be at the moment.
\item Pretend you've decided to see this movie tonight and you're taking a friend.
\item Use the internet to look up the details of the city, the theater name, showtimes offered, ticket prices, and any additional options like 3D, etc.
\item MAIN TASK: Pretend you call your personal assistant on the phone who will book the ticket for you.  Write the conversation that would happen between you and your assistant in order to buy two tickets. 
\item MAKE SURE the assistant asks about all relevant details (see \#4) INCLUDING the number of tickets needed. BUT you should choose the order that makes sense to you as far what details to ask (theater, times, etc)
\item You can assume you already have an account with the ticket service--so no credit card information is necessary.
\item The assistant should end the conversation by confirming that your tickets are being sent to your mobile device as a text message. (And you can respond thanks, goodbye, ok, etc. for a final closing turn, if you like).
\end{enumerate}
\begin{itemize}
\item {\bf YOUR TASK}: Write the conversation that results between you and your assistant. It must be at least 10 turns long (for both you and the assistant).  Below we have provided 15 turns in case you need more. KEEP IT NEW AND FRESH! DON'T REPEAT DIALOGUES FROM THE PAST!
\end{itemize}
\caption{Sample instructions for written ``self-dialogs"}
\label{fig:self_ins}
\end{figure}

\begin{figure}[!ht]
\small
\begin{tabular}{rp{5.0cm}}
USER: &Hi I would like to buy 2 tickets for Shazam!\\
ASSISTANT: &What city would you like to see this movie?\\
USER: &Ontario, California\\
ASSISTANT: &Ok, I'll check that location for you.\\
USER: &I would prefer the Edwards Ontario Mountain Village, since it's closest to me and my guest.\\
ASSISTANT: &What time is best for you?\\
USER: &Either 4 or 6 pm.\\
ASSISTANT: &I'm sorry, but it looks like the 4:10 and the 6:10 pm showings are sold out.\\
USER: &That's too bad. I really wanted to see that movie.\\
ASSISTANT: &I'm sorry. Is there another movie you would like to see?\\
USER: &How about Captain Marvel at the Edwards Ontario Mountain theater.\\
ASSISTANT: &Show times are 3:45, 7:10 and 10:10 pm.
Which would you like?\\
USER: &I am interested in the 7:10 showing.\\
ASSISTANT: &I'm sorry, it looks like the 7:10 showing is also sold out.\\
USER: &Wow, that's too bad.\\
ASSISTANT: &I'm sorry. Is there another movie you would like me to look up?\\
USER: &No, I think I'll pass on the movies tonight since those were the two I really wanted to see.\\	ASSISTANT: &If you want, I can check another theater.\\
USER: &No, that's fine.  Thank you for your help.\\
ASSISTANT: &You're welcome.\\
\end{tabular}
\caption{Sample one-person, written dialog}
\label{fig:self-example}
\end{figure}

\subsection{Self-dialogs (one-person written dataset)}
\label{sec:self-dialogs}
While the two-person approach to data collection creates a realistic scenario for robust, spoken dialog data collection, this technique is time consuming, complex and expensive, requiring considerable technical implementation as well as administrative procedures to train and manage agents and crowdsourced workers. In order to extend the Taskmaster dataset at minimal cost, we use an alternative self-dialog approach in which crowdsourced workers write the full dialogs themselves (i.e. interpreting the roles of both user and assistant).

\subsubsection{Task scenarios and instructions}
Targeting the same six tasks used for the two-person dialogs, we again engaged the Amazon Mechanical Turk worker pool to create self-dialogs, this time as a written exercise. In this case, users are asked to pretend they have a personal assistant who can help them take care of various tasks in real time. They are told to imagine a scenario in which they are speaking to their assistant on the phone while the assistant accesses the services for one of the given tasks. They then write down the entire conversation. Figure \ref{fig:self_ins} shows a sample set of instructions.

\subsubsection{Pros and cons of self-dialogs} \label{section: pros_and_cons_of_ self_dialogs}
The self-dialog technique renders quality data and avoids some of the challenges seen with the two-person approach. To begin, since the same person is writing both sides of the conversation, we never see misunderstandings that lead to frustration as is sometimes experienced between interlocutors in the two-person approach. In addition, all the self-dialogs follow a reasonable path even when the user is constructing conversations that include understanding errors or other types of dialog glitches such as when a particular choice is not available. As it turns out, crowdsourced workers are quite effective at recreating various types of interactions, both error-free and those containing various forms of linguistic repair. The sample dialog in Figure \ref{fig:self-example} shows the result of a self-dialog exercise in which workers were told to write a conversation with various ticket availability issues that is ultimately unsuccessful.

Two more benefits of the self-dialog approach are its efficiency and cost effectiveness. We were able to gather thousands of dialogs in just days without transcription or trained agents, and spent roughly six times less per dialog. Despite these advantages, the self-dialog written technique cannot recreate the disfluencies and other more complex error patterns that occur in the two-person spoken dialogs which are important for model accuracy and coverage.

\subsection{Annotation}
\label{sec:annotation}
We chose a highly simplified annotation approach for Taskmaster-1 as compared to traditional, detailed strategies which require robust agreement among workers and usually include dialog state and slot information, among other possible labels. Instead we focus solely on API arguments for each type of conversation, meaning just the variables required to execute the transaction. For example, in dialogs about setting up UBER rides, we label the ``to" and ``from" locations along with the car type (UberX, XL, Pool, etc). For movie tickets, we label the movie name, theater, time, number of tickets, and sometimes screening type (e.g. 3D vs. standard). A complete list of labels is included with the corpus release.

As discussed in Section \ref{section:agents}, to encourage diversity, at times we explicitly ask users to change their mind in the middle of the conversation, and the agents to tell the user that the requested item is not available. This results in conversations having multiple instances of the same argument type. To handle this ambiguity, in addition to the labels mentioned above, the convention of either ``accept” or ``reject" was added to all labels used to execute the transaction, depending on whether or not that transaction was successful.

\begin{figure}
\small
\begin{tabular}{rp{5.0cm}}
USER: &Finally, I need the table to be for three people and 8pm.\\
ASSISTANT: &One moment....OK, I have your table for three ({\bf num.guests.accept}) at 8pm ({\bf time.reservation.accept}) reserved.\\
\end{tabular}
\caption{Indicating transaction status with ``accept" or ``reject"}
\label{fig:label-example}
\end{figure}

In Figure \ref{fig:label-example}, both the number of people and the time variables in the assistant utterance would have the ``.accept" label indicating the transaction was completed successfully. If the utterance describing a transaction does not include the variables by name, the whole sentence is marked with the dialog type. For example, a statement such as \emph{The table has been booked for you} would be labeled as {\bf reservation.accept}.

\section{Dataset Analysis}

\begin{table}
\begin{tabular}{lcc}
\toprule
Statistic              & Self-dialogs & Two Person  \\ \midrule
\# unique words         & 17,275            & 13,490\\
\# utterances           & 110,074           & 132,407\\
\# dialogs              & 5000              & 5000          \\
Avg. utterances         & 22.01             & 24.04\\
per dialog & & \\
Avg. tokens             & 8.62              & 7.54 \\
per utterance & & \\
Perplexity              & 16.28            & 6.44\\
BLEU                    & 4.73             & 15.16\\ 
Joint-Perplexity        & 16.44            & 6.04\\
Joint-BLEU              & 5.80             & 13.09\\ \bottomrule
\end{tabular}
\caption{Statistics comparison: Self-dialogs vs 
two person corpus both containing 5k dialogs. Perplexity and BLEU are reported for Transformer baseline. Joint-Perplexity and Joint-BLEU are perplexity/BLEU scores from the joint training of self-dialogs and two-person but evaluated with their respective test sets.}
\label{table2: one_person__two_person}
\end{table}

\subsection{Self-dialogs vs MultiWOZ}

We quantitatively compare our self-dialogs (Section \ref{sec:self-dialogs})  with the MultiWOZ dataset in Table \ref{table1: one_person__multiwoz}. Compared to MultiWOZ, we do not ask the users and assistants to stick to detailed scripts and do not restrict them to have conversations surrounding a small knowledge base. Table \ref{table1: one_person__multiwoz} shows that our dataset has more unique words, and has almost twice the number of utterances per dialog than the MultiWOZ corpus.
Finally, when trained with the Transformer \cite{transformer} model, we observe significantly higher perplexities and lower BLEU scores for our dataset compared to MultiWOZ suggesting that our dataset conversations are difficult to model.  Finally, Table \ref{table1: one_person__multiwoz} also shows that our dataset contains close to 10 times more real-world named entities than MultiWOZ and thus, could potentially serve as a realistic baseline when designing goal oriented dialog systems. MultiWOZ has only 1338 unique named entities and only 4510 unique values (including date, time etc.) in their datatset. 

\subsection{Self-dialogs vs Two-person}
In this section, we quantitatively compare 5k conversations each of self-dialogs (Section \ref{sec:self-dialogs}) and two-person (Section \ref{sec:woz}).
From Table \ref{table2: one_person__two_person}, we find that self-dialogs exhibit higher perplexity ( almost 3 times) compared to the two-person conversations suggesting that self-dialogs are more diverse and contains more non-conventional conversational flows which is inline with the observations in Section-\ref{section: pros_and_cons_of_ self_dialogs}. 
While the number of unique words are higher in the case of self-dialogs, conversations are longer in the two-person conversations. 
We also report metrics by training a single model on both the datasets together.

\subsection{Baseline Experiments: Response Generation}
We evaluate various seq2seq architectures \cite{seq2seq} on our self-dialog corpus using both automatic evaluation metrics and human judgments. 
Following the recent line of work on generative dialog systems \cite{conv-seq2seq}, we treat the problem of response generation given the dialog history as a conditional language modeling problem. Specifically we want to learn a conditional probability distribution $P_{\theta}(U_{t}|U_{1:t-1})$ where $U_{t}$ is the next response given dialog history $U_{1:t-1}$. Each utterance $U_i$ itself is comprised of a sequence of words $w_{i_1}, w_{i_2} \ldots w_{i_k}$. The overall conditional probability is factorized autoregressively as $$P_{\theta}(U_{t}|U_{1:t-1}) = \prod_{i=1}^{n} P_{\theta}(w_{t_i}|w_{t_{1:i-1}},U_{1:t-1})$$

$P_{\theta}$, in this work, is parameterized by a recurrent, convolution or Transformer-based seq2seq model.

\textbf{n-gram}: We consider 3-gram and 4-gram conditional language model baseline with interpolation. We use random grid search for the best coefficients for the interpolated model. 

\textbf{Convolution}: We use the \textit{fconv} architecture \citep{fconv2017} and default hyperparameters from the fairseq \citep{ott2019fairseq} framework.\footnote{https://github.com/pytorch/fairseq} We train the network with ADAM optimizer \cite{adam} with learning rate of 0.25 and  dropout probability set to 0.2.

\textbf{LSTM}: We consider LSTM models \cite{lstm} with and without attention \cite{attention} and use the tensor2tensor \citep{tensor2tensor} framework for the LSTM baselines. We use a two-layer LSTM network for both the encoder and the decoder with 128 dimensional hidden vectors.

\textbf{Transformer}: As with LSTMs, we use the tensor2tensor framework for the Transformer model. 
Our Transformer \citep{transformer} model uses 256 dimensions for both input embedding and hidden state, 2 layers and 4 attention heads.
For both LSTMs and Transformer, we train the model with ADAM optimizer ($\beta_{1} = 0.85$, $\beta_{2} = 0.997$) and dropout probability set to 0.2.

\textbf{GPT-2}: Apart from supervised seq2seq models, we also include results from pre-trained GPT-2 \cite{open} containing 117M parameters.

\begin{table}
\small
\begin{tabular}{lccccl}
\toprule
Baseline & PPL   & BLEU & Ratings & Rank  \\
Models & & & (LIKERT) &\\
\midrule
GPT-2 (117M)       & - & 0.26  & - & - \\ 
\midrule
3-gram             & 38.12 & 0.20 & -  & -\\
4-gram             & 34.49 & 0.21 & -  & -\\
LSTM               & 25.73 & 4.45 & -  & -\\
Convolution        & 21.25 & 5.09 & 2.89  & 3\\
LSTM-attention     & 20.05 & 5.12 & \textbf{3.51}  & 2\\
Transformer       & \textbf{18.19} & \textbf{6.11} & 3.22  & \textbf{1}\\ \bottomrule
\end{tabular}
\caption{Evaluation of various seq2seq architectures \cite{seq2seq} on our self-dialog corpus using both automatic evaluation metrics and human judgments. Human evaluation ratings in the 1-5 LIKERT scale (higher the better), and human ranking are averaged over 500 x 3 ratings (3 crowdsourced workers per rating). }
\label{table3: one_person_models}
\end{table}

We evaluate all the models with perplexity and BLEU scores (Table \ref{table3: one_person_models}).
Additionally, we perform two kinds of human evaluation - Ranking and Rating (LIKERT scale) for the top-3 performing models - Convolution, LSTM-attention and Transformer.
For the ranking task, we randomly show 500 partial dialogs and generated responses of the top-3 models from the test set to three different crowdsourced workers and ask them to rank the responses based on their relevance to the dialog history.
For the rating task, we show the model responses individually
to three different crowdsourced workers and ask them to rate the responses on a 1-5 LIKERT scale based on their appropriateness to the dialog history. 
From Table-\ref{table4: inter-annotator}, we see that inter-annotator reliability scores (Krippendorf’s Alpha) are higher for the ranking task compared to the rating task.
From Table \ref{table3: one_person_models}, we see that Transformer is the best performing model on automatic evaluation metrics. It is interesting to note that there is a strong correlation between BLEU score and human ranking judgments.

\begin{table}
\small
\begin{tabular}{lc}
\toprule
Evalation  & Inter-Annotator Reliability\\
method & (Krippendorf’s Alpha)
\\
\midrule
Rating (1-5 LIKERT) & 0.21\\
Ranking         & 0.29\\ \bottomrule
\end{tabular}
\caption{Inter-Annotator Reliability scores of seq2seq model responses computed for 500 self-dialogs from the test set, each annotated by 3 crowdsourced workers.}
\label{table4: inter-annotator}
\end{table}

\subsection{Baseline Experiments: Argument Prediction}
Next, we discuss a set of baseline experiments for the task of argument prediction. API arguments are annotated as spans in the dialog (Section \ref{sec:annotation}).
We formulate this problem as mapping text conversation to a sequence of output arguments. Apart from the seq2seq Transformer baseline, we consider an additional model - an enhanced Transformer seq2seq model where the decoder can choose to copy  from the input or generate from the vocabulary \cite{copy,copy-1}. Since all the API arguments are input spans, the copy model having the correct inductive bias achieves the best performance.

\begin{table}
\small
\begin{tabular}{lc}
\toprule
Model & Micro F1 (\%)\\
\midrule
Transformer  & 48.73\\
Transformer + copy  & 51.79 \\
\bottomrule
\end{tabular}
\caption{API Argument prediction accuracy for Self-dialogs. API arguments are annotated as spans in the utterances.}
\label{table5: annotation}
\end{table}

\section{Conclusion}
To address the lack of quality corpora for data-driven dialog system research and development, this paper introduces  Taskmaster-1, a dataset that provides richer and more diverse language as compared to current benchmarks since it is based on unrestricted, task-oriented conversations involving more real-word entities. In addition, we present two data collection methodologies, both spoken and written, that ensure both speaker diversity and conversational accuracy. Our straightforward, API-oriented annotation technique is much easier for annotators to learn and simpler to apply. We give several baseline models including state-of-the-art neural seq2seq architectures, provide qualitative human performance evaluations for these models, and find that automatic evaluation metrics correlate well with human judgments. 
\bibliography{references}
\bibliographystyle{acl_natbib}
\end{document}